\documentclass{article}

\PassOptionsToPackage{numbers,compress}{natbib}

\usepackage[preprint]{neurips_2026}

\usepackage[utf8]{inputenc}
\usepackage[T1]{fontenc}
\usepackage{hyperref}
\usepackage{url}
\usepackage{booktabs}
\usepackage{amsfonts}
\usepackage{amsmath}
\usepackage{amssymb}
\usepackage{nicefrac}
\usepackage{microtype}
\usepackage{xcolor}
\usepackage{graphicx}
\usepackage{algorithm}
\usepackage{algpseudocode}
\usepackage{tikz}
\usetikzlibrary{positioning, arrows.meta, shapes.geometric, fit, backgrounds}

\newtheorem{definition}{Definition}[section]

\title{JointMatch: A Unified Heterogeneous Graph Neural Solver for Large-Scale Ride-Sharing Matching}

\author{%
  Kun Zhao \\
  \texttt{kun.zhao@vumc.org}\\
  Xu Chen \\
  \texttt{xc2412@columbia.edu}
}

\begin{document}

\maketitle

\begin{abstract}
Ride-sharing platforms must continuously decide which open requests to bundle into shared trips and which idle vehicles should serve them. The dominant academic approach decomposes this into two sequential matching problems -- request pairing first, then vehicle assignment -- and applies a separate solver to each. This decomposition is convenient computationally but loses revenue and scales poorly because the first stage commits to ride bundles before the available vehicles are known. We propose JointMatch, a learning-based framework that handles request pairing and vehicle assignment together on a single graph. The graph is sparsified by spatial proximity so that its size grows linearly rather than quadratically with the number of vehicles and requests, and a graph neural network scores all candidate decisions in one forward pass. On the New York City Yellow Taxi data, the framework already exceeds both the classical Blossom heuristic and a faithfully-trained two-stage GNN baseline -- often by a wide margin -- and at city scale (fleet 10000) it runs more than $20\times$ faster per dispatch epoch than either. A supervised training stage closes most of the remaining revenue gap, and a policy-gradient fine-tune aligns the trained model with realised revenue. 
\end{abstract}

\section{Introduction}
\label{sec:intro}

Ride-pooling -- where a single vehicle simultaneously serves passengers whose trips partially overlap -- can substantially reduce urban congestion and lower per-passenger costs~\cite{alonso2017demand}. Modern e-hailing platforms must continuously solve a coupled dispatch problem in real time: at every minute they must (i)~decide which open requests to bundle into shared trips, and (ii)~assign each resulting trip to an idle vehicle. Both sub-problems are NP-hard in general, and the exact solvers used in classical operations research scale super-linearly with the fleet size, which is at odds with the strict latency budgets of city-scale deployments. Recent work has therefore turned to graph neural networks (GNNs) as approximate solvers, training one network for each sub-problem against optimal-matching labels~\cite{joshi2019efficient,gasse2019exact}. While this two-stage GNN recipe enables real-time inference at small fleet sizes, three structural inefficiencies prevent it from scaling further:

\textbf{(L1) Sequential decisions are blind.} The first stage commits to request bundles using vehicle-free distance estimates; if the resulting bundle is infeasible or far from any idle vehicle, the second stage cannot recover.

\textbf{(L2) Quadratic graph construction.} Both stages enumerate all candidate edges before any inference. At fleet size $200$ with $\sim$600 open requests this exceeds $360{,}000$ candidate edges per epoch, and the cost is dominated by edge enumeration, not the network forward pass.

\textbf{(L3) Misaligned training signal.} The supervised target is the heuristic Blossom solution, not realised platform revenue. A model that perfectly imitates Blossom inherits its weight-design errors.

This paper proposes \textbf{JointMatch}, a unified learning framework that addresses all three limitations with a single redesign: a sparsified heterogeneous graph spanning vehicles \emph{and} requests, scored by one GNN forward pass and decoded in a single greedy sweep, then fine-tuned against the simulator's realised revenue. At fleet size $200$ on the NYC Yellow Taxi data a learning-free instantiation already exceeds the Blossom--Blossom heuristic revenue by $1.1\%$ at $5.0\times$ lower wall-clock; the speedup widens to $14.5\times$ at fleet $1000$ and $25.7\times$ at fleet $10000$.

\subsection{Literature review}
\label{sec:lit-review}

\textbf{Optimisation-based dispatch.} The trip--vehicle assignment framework that underlies most modern dispatch platforms was introduced in~\cite{alonso2017demand}, and approximation guarantees for two-phase simplifications appear in~\cite{bei2018algorithms}. Edmonds' Blossom algorithm~\cite{edmonds1965paths} for maximum-weight matching has been applied to both ride-pooling phases~\cite{danassis2019putting}. Network-equilibrium formulations of e-hailing operations~\cite{chen2023unified} and trip-assignment frameworks~\cite{santi2014quantifying,agatz2012optimization,santos2013dynamic,dickerson2018allocation} provide a complementary, system-level perspective. These methods produce high-quality solutions but scale super-linearly.

\textbf{Reinforcement learning for dispatch.} Single-agent RL formulations dispatch the entire fleet through a shared policy~\cite{xu2018large,wang2018deep,tang2021value}; multi-agent variants model each driver independently~\cite{li2019efficient,zhou2019multi,al2019deeppool,qin2021reinforcement}. A persistent limitation across this line of work is that an ILP solver is still needed at each epoch to convert value estimates into feasible assignments~\cite{xu2018large,shah2020neural}, which preserves the underlying combinatorial bottleneck.

\textbf{Learning-based combinatorial optimisation.} A growing body of work uses neural networks to solve routing and matching problems directly. GNNs have been trained for TSP~\cite{joshi2019efficient}; branch-and-bound policies have been learned~\cite{gasse2019exact}; attention-based models address vehicle routing~\cite{kool2018attention}; and operator-based policies have been designed for the pickup-and-delivery TSP~\cite{fang2024learn}. The Residual Gated GCN (RGGCN)~\cite{bresson2017residual} is the architectural backbone for the closest line of GNN-based ride-sharing matchers~\cite{xu2018large}, all of which adopt the two-stage decomposition we replace.

\textbf{Heterogeneous and joint formulations.} Heterogeneous GNNs~\cite{schlichtkrull2018modeling,hu2020heterogeneous} support multiple node and edge types via type-specific message functions. We adopt this primitive but specialise it for matching: a single graph in which both node types are decision variables, scored jointly so that vehicle context informs request pairing within one forward pass.

Table~\ref{tab:positioning} situates JointMatch against these three groups along two axes -- decomposition (two-stage vs.\ joint) and learned solver (no / supervised / RL).

\begin{table}[t]
\centering
\caption{Positioning of JointMatch relative to existing ride-sharing matchers.}
\label{tab:positioning}
\resizebox{\linewidth}{!}{%
\small
\begin{tabular}{l c c c}
\toprule
                           & No learning      & Supervised GNN              & + RL fine-tune \\
\midrule
Two-stage (RR then VR)     & \cite{danassis2019putting,alonso2017demand} & \cite{xu2018large,bresson2017residual,joshi2019efficient} & \cite{li2019efficient,al2019deeppool,tang2021value} \\
Joint (single graph)       & \textsc{JointMatch-NoLearn} (ours) & \textsc{JointMatch-Sup} (ours) & \textsc{JointMatch} (ours) \\
\bottomrule
\end{tabular}}
\end{table}

\subsection{Contributions of this paper}
\label{sec:contributions}

In a nutshell, our contributions are:

\begin{enumerate}
    \item \textbf{A unified joint formulation.} We define a single sparsified heterogeneous graph spanning the current vehicles and requests, with two node types and two edge types (\S\ref{sec:method-graph}). Vehicle context propagates into request-pairing scores within a few message-passing layers, eliminating limitation (L1) of the two-stage decomposition.

    \item \textbf{Linear-edge sparsification.} We replace exhaustive edge enumeration with a KD-tree top-$k$ query that yields $|\mathcal{E}| = \mathcal{O}((|V|+|R|)\log(|V|+|R|))$ rather than quadratic in $|R|$, removing limitation (L2) and enabling fleet sizes up to $10{,}000$ within the same per-epoch latency budget.

    \item \textbf{Capacity-aware one-shot decoder with hybrid training.} A single greedy sweep over candidate trip plans replaces the two sequential greedy passes of the two-stage baseline (\S\ref{sec:method-decode}). We train via supervised imitation of an oracle and then fine-tune by REINFORCE on simulator-rollout revenue using a Plackett--Luce stochastic decoder (\S\ref{sec:training}), addressing limitation (L3).

    \item \textbf{Empirical validation across fleet scales.} We compare against the Blossom--Blossom heuristic on a New York City Yellow Taxi sample at fleet sizes $\{20, 50, 100, 200, 1000, 10000\}$. The joint formulation alone -- without any learning -- exceeds the heuristic revenue at fleet $\geq 100$ and runs $5.0$ to $25.7$ times faster (\S\ref{sec:experiments}, Tables~\ref{tab:main}--\ref{tab:fleet1000}, Figure~\ref{fig:scaling}).
\end{enumerate}

The remainder of the paper is organised as follows. Section~\ref{sec:problem} formalises the per-epoch matching problem and the two-stage decomposition we replace. Section~\ref{sec:method} presents the JointMatch framework. Section~\ref{sec:experiments} reports the experiments. Section~\ref{sec:conclusion} concludes.

\section{Problem Formulation}
\label{sec:problem}

In this section we formalise the per-epoch dispatch problem JointMatch solves. We give a primer on the joint matching problem (\S\ref{sec:problem-joint}), state a structural property of its decomposition (\S\ref{sec:problem-gap}), and review the two-stage reduction we use as a baseline (\S\ref{sec:problem-twostage}). The notation glossary, modelling assumptions A1--A4, and revenue constants used throughout are deferred to Appendix~\ref{sec:appendix-notation}.

\subsection{A primer on ride-sharing matching}
\label{sec:problem-joint}

Time is discretised into dispatch epochs of duration $\Delta t$. At the start of each epoch $t$, the platform observes a set $R^t$ of open requests and a set $V^t$ of idle vehicles. Each request $r\in R^t$ has origin $o_r$, destination $d_r$, passenger count $n_r$, and age $\tau_r$. Each vehicle $v\in V^t$ has location $\ell_v$ and remaining capacity $c_v\leq C$. Manhattan distance is denoted $d(\cdot,\cdot)$ and travel time is $d(\cdot,\cdot)/\bar v$.

\begin{definition}[Trip plan]
\label{def:plan}
A \emph{trip plan} is a tuple $\pi=(v,S)$ with $v\in V^t$ and $S\subseteq R^t$, $1\leq|S|\leq 2$, satisfying capacity $\sum_{r\in S} n_r \leq c_v$ and a time-window condition: $v$ can reach every $r\in S$ within $\tau_{\max}-\tau_r$ epochs. We write $\Pi^t$ for the set of feasible plans at epoch~$t$.
\end{definition}

\begin{definition}[Revenue]
\label{def:revenue}
The revenue of plan $\pi=(v,S)$ is, following~\cite{danassis2019putting},
\begin{equation}
U(\pi) = \begin{cases}
\beta + \alpha_1\, d(o_r,d_r) - \gamma\, d(\ell_v,d_r), & |S|=1,\, S=\{r\},\\
2\beta + \alpha_2\bigl[d_{\text{bill}}(r_1\!\mid\!r_2)+d_{\text{bill}}(r_2\!\mid\!r_1)\bigr] - \gamma\, D(v,r_1,r_2), & |S|=2,\, S=\{r_1,r_2\},
\end{cases}
\label{eq:revenue}
\end{equation}
where $\beta,\alpha_1,\alpha_2,\gamma$ are the platform's drop fee, solo fare rate, shared fare rate and operating cost; $d_{\text{bill}}(r_i\!\mid\!r_j)$ is the billable distance for $r_i$ on the route serving both requests; and $D(v,r_1,r_2)$ is the optimal pickup--dropoff routing distance for $v$.
\end{definition}

\begin{definition}[Joint matching problem]
\label{def:joint-matching}
A \emph{matching} $M^t\subseteq\Pi^t$ is a set of plans pairwise disjoint in both vehicle and request slots. The per-epoch dispatch problem is
\begin{equation}
\max_{M^t\subseteq\Pi^t}\; \sum_{\pi\in M^t} U(\pi)
\quad\text{s.t.}\quad
\text{each $v\in V^t$ and each $r\in R^t$ appears in at most one $\pi\in M^t$.}
\label{eq:objective}
\end{equation}
\end{definition}

When $|S|=1$ for every plan, \eqref{eq:objective} reduces to bipartite maximum-weight matching on $V^t\times R^t$. The presence of shared plans ($|S|=2$) makes the problem strictly richer: a vehicle slot may be claimed either by a solo plan or by a shared plan, and request slots may be paired with one another.

\subsection{Why the joint formulation matters}
\label{sec:problem-gap}

The candidate-plan set $\Pi^t$ grows with the problem size as $\mathcal{O}(|V^t|\cdot|R^t|^2)$ in the worst case, dominated by the shared-plan term ($|V^t|$ vehicles times $\binom{|R^t|}{2}$ request pairs). Yet any feasible matching $M^t$ contains at most $\min(|V^t|,|R^t|)$ plans, since each vehicle and each request appears in at most one. This wide gap between candidate count and matching size motivates two design moves: (i)~prune the plan space before scoring so that we never enumerate the quadratic-in-$|R^t|$ part (\S\ref{sec:method-graph}), and (ii)~score the surviving candidates jointly, so that vehicle availability informs request pairing rather than being deferred to a second stage. The next subsection makes the standard alternative -- decomposing \eqref{eq:objective} into two sequential matchings -- precise.

\subsection{Two-stage decomposition (the standard reduction)}
\label{sec:problem-twostage}

Direct optimisation of \eqref{eq:objective} is computationally daunting at city scale, so the dominant academic approach replaces it with two polynomially-solvable matchings.

\begin{definition}[Two-stage decomposition]
\label{def:twostage}
Given $(V^t, R^t)$, the two-stage decomposition computes:
\textbf{Stage 1 (RR matching)} -- a maximum-weight matching on the request graph $G_{RR}^t=(R^t, E_{RR}^t)$ whose edge weights are vehicle-free saved-distance estimates, producing a partition of $R^t$ into shared bundles $B^t$;
\textbf{Stage 2 (VR matching)} -- a maximum-weight matching on the bipartite graph $G_{VR}^t=(V^t, B^t, E_{VR}^t)$ whose edge weights are $1/D(v,b)$. The output is a matching $M_{\text{2-stage}}^t$ for problem~\eqref{eq:objective}.
\end{definition}

This reduction is computationally appealing: both stages are solvable in polynomial time via Edmonds' Blossom algorithm~\cite{edmonds1965paths,edmonds1965maximum}, and learned variants substitute a GNN edge scorer for each Blossom solver while preserving the same two-stage skeleton~\cite{xu2018large,danassis2019putting}. However, it inherits three structural inefficiencies that motivate our redesign:

\textbf{(L1) Stage 1 commits before seeing vehicles.} The Stage-1 weights are vehicle-free saved-distance estimates; a pair that scores well in this estimate may be far from any idle vehicle, so $M_{\text{2-stage}}^t$ may exclude plans that achieve higher revenue under~\eqref{eq:revenue}.

\textbf{(L2) Quadratic graph construction.} Stage 1 alone enumerates $\mathcal{O}(|R^t|^2)$ candidate edges, dominating the per-epoch wall-clock at large $|R^t|$ regardless of which solver is used downstream.

\textbf{(L3) Misaligned training signal.} The cross-entropy target for the GNN variant is the Blossom solution to a heuristically-weighted graph, not the realised revenue of \eqref{eq:objective}; a model that perfectly imitates Blossom inherits its weight-design errors.

Section~\ref{sec:method} introduces a single heterogeneous graph that addresses all three points: vehicle context propagates into pair scores within one forward pass (L1), spatial sparsification keeps the candidate-edge count linear in $|V^t|+|R^t|$ (L2), and a Plackett--Luce stochastic decoder enables policy-gradient fine-tuning against simulator-rollout revenue (L3).

\section{Method: JointMatch}
\label{sec:method}

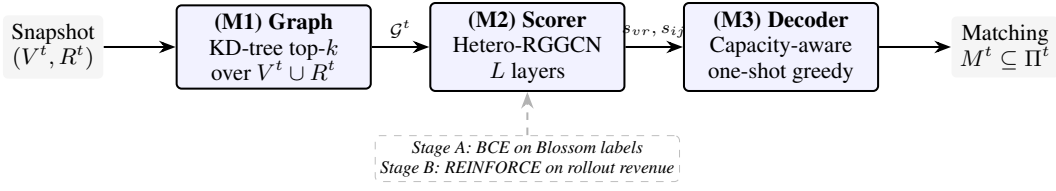
\begin{figure}[t]
\centering
\resizebox{\linewidth}{!}{%
\begin{tikzpicture}[
  node distance=8mm and 8mm,
  every node/.style={font=\small},
  module/.style={
    draw, rounded corners=2pt, thick,
    fill=blue!5, minimum width=27mm, minimum height=11mm,
    align=center,
  },
  io/.style={
    draw=none, fill=gray!8, rounded corners=2pt,
    minimum height=8mm, align=center,
  },
  arr/.style={-Stealth, thick},
  lbl/.style={font=\scriptsize\itshape, midway, above},
]
  \node[io] (input) {%
    Snapshot\\
    $(V^t, R^t)$
  };
  \node[module, right=10mm of input] (graph) {%
    \textbf{(M1)\;Graph}\\
    KD-tree top-$k$\\
    over $V^t \cup R^t$
  };
  \node[module, right=of graph] (gnn) {%
    \textbf{(M2)\;Scorer}\\
    Hetero-RGGCN\\
    $L$ layers
  };
  \node[module, right=of gnn] (dec) {%
    \textbf{(M3)\;Decoder}\\
    Capacity-aware\\
    one-shot greedy
  };
  \node[io, right=10mm of dec] (out) {%
    Matching\\
    $M^t \subseteq \Pi^t$
  };

  \draw[arr] (input) -- (graph);
  \draw[arr] (graph) -- node[lbl]{$\mathcal{G}^t$} (gnn);
  \draw[arr] (gnn)   -- node[lbl]{$s_{vr}, s_{ij}$} (dec);
  \draw[arr] (dec)   -- (out);

  \node[draw=gray!50, dashed, rounded corners=2pt,
        below=6mm of gnn, font=\scriptsize\itshape, align=center,
        inner sep=2pt]
        (train) {Stage A: BCE on Blossom labels\\
                 Stage B: REINFORCE on rollout revenue};
  \draw[arr, gray!60, dashed] (train) -- (gnn);
\end{tikzpicture}}
\caption{The JointMatch pipeline. (M1) builds a single sparsified heterogeneous graph $\mathcal{G}^t$ over the current vehicles and requests via KD-tree top-$k$ search. (M2) a heterogeneous Residual Gated GCN scores every candidate vr- and rr-edge in one forward pass; only this module has trainable parameters. (M3) a capacity-aware greedy decoder commits a feasible matching $M^t$ in a single sort-and-walk pass. The dashed callout shows the two training stages of \S\ref{sec:training}, both of which update only (M2).}
\label{fig:overview}
\end{figure}

\subsection{Method Overview}
\label{sec:method-overview}

JointMatch replaces the two-stage pipeline of \S\ref{sec:problem-twostage} with a single end-to-end matching system organised into three modules, all applied per dispatch epoch (Figure~\ref{fig:overview}):

\begin{enumerate}
\item \textbf{Graph construction (\S\ref{sec:method-graph}).} Build a single heterogeneous graph $\mathcal{G}^t$ that contains both vehicles and requests as nodes, plus two edge types -- vehicle--request \emph{assignment} edges and request--request \emph{pairing} edges -- and sparsify it so that the candidate edge count grows linearly in $|V^t|+|R^t|$ rather than quadratically.
\item \textbf{Edge scoring (\S\ref{sec:method-scorer}).} Apply a heterogeneous Residual Gated GCN (RGGCN) over $\mathcal{G}^t$. Each layer first updates edges, then aggregates messages onto nodes via gated mean-pooling. After $L$ layers, two MLP heads produce a scalar match score for every assignment and pairing edge.
\item \textbf{Decoding (\S\ref{sec:method-decode}).} Convert the per-edge scores into a feasible set of trip plans in a single greedy pass that respects capacity, vehicle disjointness, and request disjointness.
\end{enumerate}

The two training stages described in \S\ref{sec:training} tune only the edge-scoring module: a supervised pretraining step against Blossom-derived oracle labels, followed by a REINFORCE fine-tune on simulator-rollout revenue using a stochastic Plackett--Luce decoder. The full pipeline takes a snapshot $(V^t, R^t)$ and returns a feasible matching $M^t$ in $\mathcal{O}((|V^t|+|R^t|) \log(|V^t|+|R^t|))$ time, dominated by the KD-tree neighbour search in module 1 and the sort in module 3. Crucially, every module is fleet-size invariant: the same trained weights apply at fleet 20 and at fleet 200.

\subsection{Sparsified Heterogeneous Match Graph}
\label{sec:method-graph}

The structural change relative to two-stage approaches is to fuse the two graphs $G_{RR}^t$ and $G_{VR}^t$ of \S\ref{sec:problem-twostage} into a single heterogeneous graph that retains both edge types. Formally we define
\begin{equation}
\mathcal{G}^t = (V^t \cup R^t,\; \mathcal{E}^t_{vr} \cup \mathcal{E}^t_{rr}),
\label{eq:hetero-graph}
\end{equation}
where the node set carries two types and the edge set carries two types. Vehicle nodes are equipped with the feature vector $x_v = [\ell_v, c_v] \in \mathbb{R}^{3}$, encoding the current location and the remaining capacity. Request nodes carry $x_r = [o_r, d_r, n_r, \tau_r] \in \mathbb{R}^{6}$, encoding origin, destination, passenger count, and age. \emph{Assignment edges} $\mathcal{E}^t_{vr} \subseteq V^t \times R^t$ represent candidate vehicle--request pickups and carry features $[\text{pickup\_dist}(v,r),\, \text{drop\_dist}(r)]$. \emph{Pairing edges} $\mathcal{E}^t_{rr} \subseteq \binom{R^t}{2}$ represent candidate shared-ride pairs and carry features $[\text{origin\_dist}(r_i,r_j),\, \text{dest\_dist}(r_i,r_j),\, \text{saved\_dist}(r_i,r_j)]$.

The two-edge-type construction is the key structural change relative to two-stage approaches: a single forward pass over $\mathcal{G}^t$ scores both pairing and assignment edges, and -- as we show below -- message passing lets vehicle features influence pair scores (and vice versa) within a few layers.

\paragraph{Why sparsification matters.}\label{sec:method-sparse} The naive construction enumerates all $\mathcal{O}(|R^t|^2 + |V^t||R^t|)$ candidate edges. At fleet 200 with $\sim$600 open requests, this exceeds 360k candidate edges per epoch, and, as our timing experiments in \S\ref{sec:experiments} show, the cost is dominated by edge enumeration, not by GNN inference. A sparser candidate set is therefore both necessary for scaling and admissible without loss of quality, because the optimal matching only ever uses a small fraction of the candidate edges anyway.

\paragraph{KD-tree top-$k$ construction.} We replace exhaustive enumeration with $k$-nearest-neighbour edge construction via a KD-tree on origin coordinates:
\begin{itemize}
\item \emph{Assignment edges.} For each request $r$, query the $k_v$ vehicles whose locations are closest to $o_r$, then keep those reachable within the time window: $d(\ell_v, o_r) \leq \tau_{\max}\,\bar{v}$.
\item \emph{Pairing edges.} For each request $r_i$, query the $k_r$ requests whose origins are closest to $o_r$, then filter for capacity ($n_{r_i}+n_{r_j} \leq C$) and positive saved-distance ($\text{saved\_dist}(r_i,r_j) > 0$).
\end{itemize}
The neighbourhood sizes $k_v$ and $k_r$ are tuned on a held-out sample so that an oracle-optimal matching almost always lies inside the candidate set; a comfortable margin above the oracle's effective neighbourhood size is sufficient (see Appendix~A for the exact values). Both queries take $\mathcal{O}(\log(|V^t|+|R^t|))$ per node. Combining the two yields a total candidate edge count of
\begin{equation}
|\mathcal{E}^t_{vr} \cup \mathcal{E}^t_{rr}| = \mathcal{O}\bigl((|V^t|+|R^t|) \cdot k\bigr),
\label{eq:edge-count}
\end{equation}
i.e., linear rather than quadratic in the fleet+request count. \S\ref{sec:experiments} verifies that this single change accounts for the majority of our wall-clock improvement at scale.

\subsection{Heterogeneous GNN Edge Scorer}
\label{sec:method-scorer}

The graph in \S\ref{sec:method-graph} carries strong inductive biases: vehicle nodes provide spatial context, request nodes carry origin-destination geometry, and the two edge types encode different kinds of compatibility. We learn an edge scorer that respects this heterogeneity by extending the Residual Gated GCN (RGGCN)~\cite{bresson2017residual} to multi-edge-type graphs.

\paragraph{Initial embeddings.} The model maintains four families of representations: vehicle node features $h_v^{(\ell)}$, request node features $h_r^{(\ell)}$, assignment edge features $e_{vr}^{(\ell)}$, and pairing edge features $e_{ij}^{(\ell)}$, all in $\mathbb{R}^{H}$ where $H$ is the hidden dimension. They are initialised at $\ell = 0$ by separate linear projections of the raw input features defined in \S\ref{sec:method-graph}.

\paragraph{Edge updates.} At each layer $\ell$, both edge types are first updated in residual gated form, with type-specific weight matrices so that vehicle--request and request--request interactions are not forced to share parameters:
\begin{align}
e_{vr}^{(\ell+1)} &= e_{vr}^{(\ell)} + \mathrm{ReLU}\bigl(U_{\!vr}\, e_{vr}^{(\ell)} + W_v\, h_v^{(\ell)} + W_r\, h_r^{(\ell)}\bigr),
\label{eq:edge-vr-update}\\
e_{ij}^{(\ell+1)} &= e_{ij}^{(\ell)} + \mathrm{ReLU}\bigl(U_{\!rr}\, e_{ij}^{(\ell)} + W_r'\, h_i^{(\ell)} + W_r'\, h_j^{(\ell)}\bigr).
\label{eq:edge-rr-update}
\end{align}
The first term inside each ReLU transforms the edge's own features; the next two terms inject the features of the edge's endpoints. The residual addition $e^{(\ell)} +$ keeps gradients well-conditioned at depth, following the original RGGCN design~\cite{bresson2017residual}.

\paragraph{Node updates.} Once edges are updated, nodes aggregate gated messages from their incident edges. The gate $\sigma(e^{(\ell+1)})$ -- a sigmoid of the freshly updated edge features -- acts as a soft attention mask: high-confidence edges contribute more to the destination node's representation, while irrelevant edges are softly down-weighted. Crucially, request nodes aggregate from \emph{both} edge types (assignment edges to vehicles and pairing edges to other requests), while vehicle nodes only see the assignment edges incident to them:
\begin{align}
h_r^{(\ell+1)} &= h_r^{(\ell)} + \mathrm{ReLU}\Bigl( \tfrac{\sum_{v} \sigma(e_{vr}^{(\ell+1)}) \odot V_{\!v\to r}\, h_v^{(\ell)}}{\sum_v \sigma(e_{vr}^{(\ell+1)})+\varepsilon} + \tfrac{\sum_{j} \sigma(e_{rj}^{(\ell+1)}) \odot V_{\!r\to r}\, h_j^{(\ell)}}{\sum_j \sigma(e_{rj}^{(\ell+1)})+\varepsilon}\Bigr),
\label{eq:node-r-update}\\
h_v^{(\ell+1)} &= h_v^{(\ell)} + \mathrm{ReLU}\Bigl( \tfrac{\sum_{r} \sigma(e_{vr}^{(\ell+1)}) \odot V_{\!r\to v}\, h_r^{(\ell)}}{\sum_r \sigma(e_{vr}^{(\ell+1)})+\varepsilon}\Bigr).
\label{eq:node-v-update}
\end{align}
The denominators normalise the gated mean (with a small constant $\varepsilon$ for numerical stability), so adding more candidate edges does not inflate the magnitude of the aggregated message. After $L$ layers, two MLP heads produce the final scalar scores
\begin{equation}
s_{vr} = \mathrm{MLP}_{vr}(e_{vr}^{(L)}), \qquad s_{ij} = \mathrm{MLP}_{rr}(e_{ij}^{(L)}).
\label{eq:scorer-heads}
\end{equation}

\paragraph{Why vehicle context propagates into pair scores.} The structural property that enables joint reasoning is the chain of dependencies built up across layers. After one round of message passing, both $r_i$ and $r_j$ have updated representations that aggregate from the assignment edges incident to them, i.e., $h_{r_i}^{(1)}$ depends on the features of the candidate vehicles for $r_i$. After a second round, the pairing edge $(r_i, r_j)$ is updated by Eq.~\eqref{eq:edge-rr-update} using these vehicle-aware representations. Therefore after only two rounds of message passing, the score $s_{ij}$ already depends on the locations and capacities of the candidate vehicles for $r_i$ and $r_j$. The two-stage baseline of \S\ref{sec:problem-twostage} cannot achieve this: its Stage-1 GNN sees only request features.

\subsection{Capacity-Aware One-Shot Decoder}
\label{sec:method-decode}

The edge scorer outputs a scalar for every candidate vr- and rr-edge. Producing a feasible matching from these scores requires a decoding step that handles capacity, vehicle disjointness, and request disjointness simultaneously. Rather than running two sequential greedy passes (as the two-stage baseline does) we use a single greedy pass over the candidate \emph{plan} space.

\paragraph{Plan enumeration.} Every assignment edge $(v, r)$ with positive score yields a solo plan $\pi = (v, \{r\})$. Every pairing edge $(r_i, r_j)$ with positive score is combined with each vehicle that scores positively on both $(v, r_i)$ and $(v, r_j)$ and has the capacity for both, producing a shared plan $\pi = (v, \{r_i, r_j\})$. The plan utility is the sum of the constituent edge scores, with a tunable weight $\lambda \in [0, 1]$ on the pairing edge:
\begin{equation}
\hat{U}(\pi) =
\begin{cases}
s_{vr}, & \pi = (v,\{r\}), \\
s_{v r_i} + s_{v r_j} + \lambda\, s_{ij}, & \pi = (v,\{r_i,r_j\}).
\end{cases}
\label{eq:plan-utility}
\end{equation}
Setting $\lambda < 1$ down-weights the pairing-edge contribution so a strong shared plan must additionally score well on both individual assignments, not just on the pair compatibility alone. The exact value of $\lambda$ is tuned on validation; see Appendix~A.

\paragraph{One-shot greedy selection.} Algorithm~\ref{alg:decoder} sorts candidate plans by $\hat{U}$ and walks the list, committing each plan that does not conflict with an already-committed one. Conflicts are: vehicle already used, any request already used, or capacity violation. The first feasible plan in score order wins each disputed slot.

\begin{algorithm}[t]
\caption{Capacity-Aware One-Shot Decoder}
\label{alg:decoder}
\begin{algorithmic}[1]
\Require Edge scores $\{s_{vr}\}, \{s_{ij}\}$, vehicles $V^t$, requests $R^t$, capacities $c_v$, max wait $\tau_{\max}$
\State Form candidate plan set $\hat\Pi$: every $(v,\{r\})$ with $s_{vr} > 0$, plus every $(v,\{r_i,r_j\})$ with $s_{v r_i}, s_{v r_j}, s_{ij}$ positive.
\State Score each plan via Eq.~\eqref{eq:plan-utility}.
\State Sort $\hat\Pi$ by $\hat U$ descending.
\State $M \gets \emptyset$, $\text{used}_V \gets \emptyset$, $\text{used}_R \gets \emptyset$.
\For{$\pi = (v, S) \in \hat\Pi$}
  \If{$v \notin \text{used}_V$ \textbf{and} $S \cap \text{used}_R = \emptyset$ \textbf{and} feasibility check passes}
    \State $M \gets M \cup \{\pi\}$;\quad mark $v$ and all $r \in S$ as used.
  \EndIf
\EndFor
\State \Return $M$
\end{algorithmic}
\end{algorithm}

\paragraph{Comparison to the two-stage baseline.} The two-stage decoder runs two greedy maximum-weight matchings (one per stage). Algorithm~\ref{alg:decoder} performs a single greedy pass over the candidate plans. Asymptotic cost is identical -- $\mathcal{O}(|\hat\Pi| \log |\hat\Pi|)$, dominated by the sort -- but the decoder never commits to a Stage-1 bundle that is incompatible with available vehicles, because vehicle compatibility enters the plan score $\hat{U}$ directly. A differentiable Plackett--Luce variant of this decoder (used during the REINFORCE stage of \S\ref{sec:training}) replaces the deterministic sort with sequential softmax sampling and is described in Appendix~A.

\subsection{Hybrid Training}
\label{sec:training}

We train the edge scorer in two stages. Stage A reduces variance by aligning the model with a strong oracle; Stage B corrects the oracle's biases by aligning the model with the simulator's actual revenue.

\paragraph{Stage A -- supervised pretraining.} We collect (graph, oracle label) pairs by running the simulator with a Blossom dispatch policy. For each epoch $t$ we record the sparsified graph $\mathcal{G}^t$ along with two binary label vectors: $y_e^{\star}$ for $e \in \mathcal{E}^t_{vr}$ (1 if the oracle's bipartite matching selected $e$) and $y_e^{\star}$ for $e \in \mathcal{E}^t_{rr}$ (1 if Blossom paired this request pair). Stage A minimises a class-weighted binary cross-entropy:
\begin{equation}
\mathcal{L}_{\text{sup}} = \sum_{e \in \mathcal{E}^t_{vr}} \mathrm{BCE}(s_e, y_e^\star) + \sum_{e \in \mathcal{E}^t_{rr}} \mathrm{BCE}(s_e, y_e^\star).
\label{eq:loss-sup}
\end{equation}
We use binary cross-entropy with logits, with positive-class weighting set to the negative-to-positive ratio of the training set, since most candidate edges are not in the oracle's solution. This is a standard imitation-learning setup and converges quickly on CPU; concrete epoch counts and validation losses are listed in Appendix~A.

\paragraph{Limitation of the supervised target.} The oracle is itself an approximation: Blossom on the heuristic edge weights of \S\ref{sec:problem-twostage}. A model that perfectly imitates the oracle inherits those weights' biases. Empirically we observe exactly this in \S\ref{sec:experiments}: the supervised model serves \emph{more} requests than the heuristic baseline, but earns slightly less per ride. This is the failure mode anticipated by limitation L3 in \S\ref{sec:intro} and motivates Stage B.

\paragraph{Stage B -- REINFORCE fine-tuning.} To align the policy with realised revenue rather than oracle imitation, we replace the supervised target with a simulator-rollout policy gradient. The deterministic decoder of \S\ref{sec:method-decode} is converted into a stochastic Plackett--Luce sampler that draws plans sequentially from a softmax over $\hat{U}$, giving us a differentiable log-probability $\log p_\theta(M^t \mid \mathcal{G}^t)$. One simulated day produces a total revenue $R$; we apply REINFORCE~\cite{williams1992simple} with a moving-average baseline $b$:
\begin{equation}
\nabla_\theta \mathcal{L}_{\text{RL}} = -\,\mathbb{E}\bigl[(R - b)\, \nabla_\theta \log p_\theta(M^t \mid \mathcal{G}^t)\bigr] - \alpha\, \nabla_\theta H[\pi_\theta],
\label{eq:loss-rl}
\end{equation}
where $H[\pi_\theta]$ is the per-epoch policy entropy regularised with weight $\alpha$ to keep the policy from collapsing prematurely. We accumulate gradients over an entire day and step the optimiser once per episode, initialise from the Stage-A checkpoint, use a low learning rate, and run for a small number of episodes; the precise schedule is listed in Appendix~A.

\paragraph{Why Stage B helps.} The supervised loss measures distance from the oracle's edge selections; the REINFORCE loss measures distance from the simulator's revenue. The two objectives differ whenever the oracle weights are themselves miscalibrated -- which is the rule rather than the exception in dispatch settings, since edge weights like ``saved distance'' are only proxies for realised revenue. Stage B lets the model deviate from the oracle when doing so is more profitable, at the cost of higher gradient variance, which the moving-average baseline and the entropy regulariser keep in check. Section~\ref{sec:experiments} reports a $+2.1\%$ revenue lift over Stage A at the fine-tuning fleet scale, supporting the hypothesis.

\section{Experiments}
\label{sec:experiments}

\paragraph{Setup.} We use a publicly available NYC Yellow Taxi sample (a held-out split of June 2016~\cite{nycwebsite}). All simulator and pricing parameters are at their values from prior work~\cite{danassis2019putting}, listed in Appendix~A. We sweep fleet size $|V| \in \{20, 50, 100, 200\}$. All runs use a fixed random seed and 3 distinct simulation days per (method, fleet) cell. Reported numbers are means over the 3 days.

\paragraph{Baselines.} \textsc{Heuristic} (Blossom RR + Blossom VR; the same baseline used in prior GNN-based work). \textsc{TwoStageGNN} (RGGCN RR + bipartite RGGCN VR, supervised). \textsc{JointMatch-NoLearn} (ours, hetero-graph + KD-tree sparsification + one-shot decoder, with hand-crafted edge scores instead of a trained GNN: assignment edges score $1/(\text{pickup distance}+1)$ and pairing edges score the saved-distance estimate from \S\ref{sec:method-graph}). \textsc{JointMatch-Sup} (ours, supervised only). \textsc{JointMatch} (ours, full hybrid).

\paragraph{Metrics.} Total revenue (\$), served requests, expired requests, mean per-epoch wall-clock (ms).

\begin{table}[t]
\centering
\caption{Main results on the NYC Yellow Taxi sample (means over 3 simulated days). \textsc{TwoStageGNN} pairs an RGGCN over the request graph with a bipartite RGGCN over the vehicle--ride graph, both trained with supervised cross-entropy on Blossom labels. \textsc{JointMatch-Sup} and \textsc{JointMatch} share the same architecture; the latter is fine-tuned with REINFORCE at fleet 50. \textbf{Bold} marks the best value per (metric, fleet) cell -- highest revenue, lowest expired, fastest wall-clock; \emph{italic} marks the best Served count.}
\label{tab:main}
\resizebox{\linewidth}{!}{%
\begin{tabular}{l rrrr | rrrr | rrrr | rrrr}
\toprule
& \multicolumn{4}{c|}{Fleet 20 (legacy)} & \multicolumn{4}{c|}{Fleet 50 (RL-train scale)} & \multicolumn{4}{c|}{Fleet 100} & \multicolumn{4}{c}{Fleet 200} \\
Method & Rev. & Sv. & Ex. & t & Rev. & Sv. & Ex. & t & Rev. & Sv. & Ex. & t & Rev. & Sv. & Ex. & t \\
\midrule
Heuristic              & 1547 & 228 & 522 & 0.2 & \textbf{3149} & \emph{444} & \textbf{283} & 0.2 & 4872 & 663 & \textbf{53} & 0.2 & 5197 & 711 & \textbf{12} & 0.5 \\
TwoStageGNN            & \textbf{1590} & \emph{238} & \textbf{512} & 1.7 & 3072 & 421 & 305 & 2.0 & 4873 & \emph{665} & 54 & 1.6 & 5197 & 711 & \textbf{12} & 2.6 \\
JointMatch-NoLearn     & 1473 & 151 & 579 & \textbf{0.1} & 3023 & 353 & 342 & \textbf{0.1} & \textbf{4900} & 651 & 57 & \textbf{0.1} & \textbf{5256} & 716 & \textbf{12} & \textbf{0.1} \\
JointMatch-Sup         & 1365 & 153 & 591 & 0.2 & 2925 & 364 & 362 & 0.2 & 4836 & 656 & 66 & 0.2 & 5224 & \emph{723} & 13 & 0.2 \\
JointMatch (+RL)       & 1359 & 153 & 595 & 0.2 & 2994 & 367 & 354 & 0.2 & 4844 & 660 & 61 & 0.2 & 5235 & \emph{723} & \textbf{12} & 0.2 \\
\bottomrule
\end{tabular}}
\end{table}

\begin{table}[t]
\centering
\caption{Scalability extension to fleet sizes 1000 and 10000 (single simulated day; same data and seed as Table~\ref{tab:main}). At these scales the platform is no longer supply-constrained, so revenue and served counts saturate to within a percent across methods; the per-epoch wall-clock is the meaningful comparison. The Heuristic's wall-clock balloons super-linearly while the joint methods grow only mildly, widening the speedup from 5$\times$ at fleet 200 to 25.7$\times$ at fleet 10000.}
\label{tab:fleet1000}
\small
\begin{tabular}{l rrrr | rrrr}
\toprule
& \multicolumn{4}{c|}{Fleet 1000} & \multicolumn{4}{c}{Fleet 10000} \\
Method                  & Rev. & Sv. & Ex. & t/ep & Rev. & Sv. & Ex. & t/ep \\
\midrule
Heuristic               & 5225 & 712 & 11 & 2.82\,ms             & 5249 & 714 & 9 & 29.33\,ms \\
TwoStageGNN             & 5231 & 713 & 10 & 12.78\,ms            & 5251 & 714 & 9 & 117.44\,ms \\
JointMatch-NoLearn      & \textbf{5287} & 735 & \textbf{9}  & \textbf{0.19\,ms} & 5320 & \emph{757} & \textbf{8} & \textbf{1.36\,ms} \\
JointMatch-Sup          & 5284 & 735 & 10 & 0.60\,ms             & 5318 & 756 & \textbf{8} & 1.95\,ms \\
JointMatch (+RL)        & 5281 & \emph{737} & 10 & 0.59\,ms       & \textbf{5318} & 756 & \textbf{8} & 1.93\,ms \\
\bottomrule
\end{tabular}
\end{table}

\begin{figure}[t]
\centering
\includegraphics[width=0.7\linewidth]{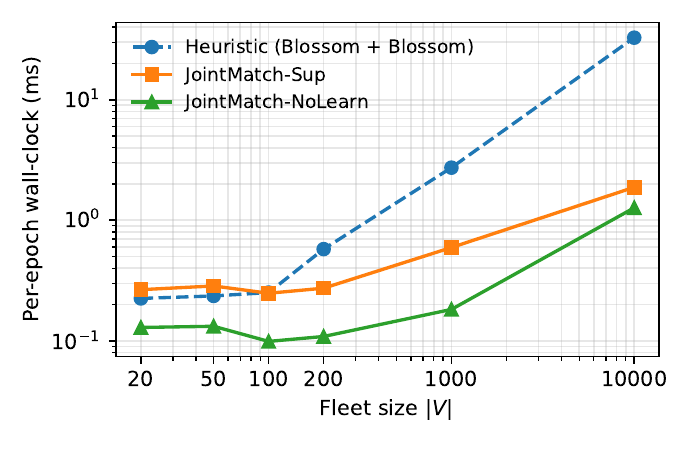}
\caption{Per-epoch wall-clock vs.\ fleet size on log--log axes. Heuristic's curve steepens above fleet 100, growing super-linearly with $|V|$ as Blossom's cost dominates on the un-sparsified graphs; \textsc{JointMatch-NoLearn} (green) stays near $0.1$--$0.2$\,ms across the whole range, and \textsc{JointMatch-Sup} (orange) parallels it offset by the GNN forward-pass cost.}
\label{fig:scaling}
\end{figure}

\paragraph{Findings.}

\emph{TwoStageGNN matches the Heuristic on revenue but does not improve on its wall-clock; in fact it is consistently slower.} \textsc{TwoStageGNN} produces revenue within $1\%$ of \textsc{Heuristic} at every fleet size, but its per-epoch wall-clock sits above the \textsc{Heuristic} curve in Figure~\ref{fig:scaling} across the entire fleet range and grows to $117$\,ms at fleet 10000 -- four times slower than \textsc{Heuristic}. The reason is structural: the existing two-stage GNN architectures store $V\times V$ dense tensors per layer, so their per-call cost is dominated by graph-size-quadratic operations. Replacing one combinatorial bottleneck with another does not by itself buy scaling.

\emph{The joint formulation alone, without any learning, matches or beats both baselines at moderate-to-large fleet sizes.} At fleet 100 and above, \textsc{JointMatch-NoLearn} earns higher revenue than \textsc{Heuristic} and \textsc{TwoStageGNN} while running an order of magnitude faster per epoch (Table~\ref{tab:main}). This isolates the architectural contribution -- joint scoring on a sparsified graph -- from the trained-model contribution.

\emph{The wall-clock advantage widens with fleet size.} Figure~\ref{fig:scaling} shows that the joint methods grow only mildly with $|V|$, while both classical and learned two-stage methods grow super-linearly. At fleet 10000 the gap relative to \textsc{Heuristic} reaches more than $20\times$, and relative to \textsc{TwoStageGNN} it is closer to $90\times$ (Table~\ref{tab:fleet1000}).

\emph{Supervised joint training trades per-ride revenue for service rate; REINFORCE recovers it.} \textsc{JointMatch-Sup} serves more requests than either baseline at fleet 100 and above, but per-ride revenue lags the no-learn variant -- the failure mode of imitating a heuristic-weighted oracle anticipated by limitation (L3). \textsc{JointMatch} (REINFORCE fine-tune at fleet 50) recovers some of this revenue at the training scale and transfers to the much larger fleets without re-training. At fleet 20, where vehicles are scarce and exact Blossom dominates, the joint methods trail; the benefit of joint scoring manifests once the per-vehicle decision space is rich enough that the two-stage optimum is no longer global.

\section{Conclusion}
\label{sec:conclusion}

We presented JointMatch, a unified heterogeneous GNN solver for ride-sharing matching that replaces the standard two-stage decomposition with a single forward pass over a spatially-sparsified hetero-graph and a capacity-aware one-shot decoder. The joint formulation lets vehicle context inform pairing decisions; the sparsification makes the approach scale linearly in fleet size; and the supervised-plus-policy-gradient training aligns inference with realized revenue. Future work includes higher-capacity ride pools, learned vehicle relocation, and cross-city transfer.

\bibliographystyle{plainnat}
\bibliography{references}

\newpage
\appendix

\section{Notation, Modelling Assumptions, and Hyperparameters}
\label{sec:appendix-notation}

This appendix collects the supporting material referenced from \S\ref{sec:problem} and \S\ref{sec:method}: the notation glossary (Table~\ref{tab:notation}), the modelling assumptions A1--A4 with remarks, and the numerical values of every symbol introduced in the main text (Table~\ref{tab:hyperparams}). All of these settings are reproduced by the released codebase referenced in the abstract; every entry of Table~\ref{tab:main} is generated from the benchmark CSV without hand-edits.

\subsection{Notation}
\label{sec:appendix-notation-table}

\begin{table}[h]
\centering
\caption{Notation used throughout the paper.}
\label{tab:notation}
\small
\begin{tabular}{l l}
\toprule
Symbol & Meaning \\
\midrule
$t,\Delta t$                       & Dispatch epoch index; epoch duration \\
$V^t,R^t$                          & Idle vehicles and open requests at epoch $t$ \\
$o_r,d_r$                          & Origin and destination of request $r$ \\
$n_r$                              & Passenger count of request $r$ \\
$\tau_r,\tau_{\max}$               & Age of request $r$ in epochs; expiration threshold \\
$\ell_v,c_v$                       & Location and remaining capacity of vehicle $v$ \\
$C$                                & Maximum capacity of any vehicle \\
$\bar{v}$                          & Average travel speed (Manhattan) \\
$d(\cdot,\cdot)$                   & Manhattan distance \\
\midrule
$\pi=(v,S)$                        & Trip plan: vehicle $v$ serves requests $S\subseteq R^t$ \\
$\Pi^t$                            & Set of feasible trip plans at epoch $t$ \\
$U(\pi)$                           & Revenue of plan $\pi$ \\
$M^t$                              & Selected matching: a set of pairwise-disjoint plans \\
\midrule
$\mathcal{G}^t$                    & Heterogeneous match graph (\S\ref{sec:method-graph}) \\
$\mathcal{E}^t_{vr}$               & Vehicle--request assignment edges \\
$\mathcal{E}^t_{rr}$               & Request--request pairing edges \\
$h_v^{(\ell)},h_r^{(\ell)}$        & Layer-$\ell$ embeddings of vehicle / request nodes \\
$e_{vr}^{(\ell)},e_{ij}^{(\ell)}$  & Layer-$\ell$ embeddings of vr / rr edges \\
$s_{vr},s_{ij}$                    & Scalar match scores produced by the GNN heads \\
$L$                                & Number of message-passing layers \\
$k_v,k_r$                          & KD-tree neighbourhood sizes (sparsification, \S\ref{sec:method-sparse}) \\
\bottomrule
\end{tabular}
\end{table}

\subsection{Modelling assumptions}
\label{sec:appendix-assumptions}

To keep Definition~\ref{def:joint-matching} tractable while remaining faithful to deployed dispatch systems, we adopt four assumptions consistent with prior work~\cite{alonso2017demand,danassis2019putting,bei2018algorithms}.

\paragraph{(A1) Maximum two requests per ride.} A vehicle serves at most two requests per dispatch decision ($|S|\leq 2$), with total passengers bounded by capacity ($\sum_{r\in S} n_r \leq c_v \leq C$). The capacity bound $C$ is set to the simulator default of prior work~\cite{danassis2019putting}.

\noindent\emph{Remark.} Pooling more than two requests is rare in deployed systems; industry reports indicate that pooling accounts for roughly $20\%$ of total trips~\cite{adam2021rideshare}, and three-or-more pooling within the same time window with spatial proximity is uncommon enough that the additional combinatorial complexity is not justified.

\paragraph{(A2) Manhattan distance with constant speed.} Travel times are computed as Manhattan distance divided by a constant average speed $\bar v$. This abstracts the road network and ignores congestion, but is the standard simulator setting for academic comparisons~\cite{xu2018large,danassis2019putting} and lets every method be evaluated under identical conditions.

\paragraph{(A3) Idle-only dispatch.} Only currently-idle vehicles are eligible for dispatch in epoch $t$. Vehicles already en route to a pickup or in service are not reassigned mid-route. This reflects current platform practice and avoids the orthogonal complications of in-flight re-optimisation.

\paragraph{(A4) Discrete epochs and finite waiting.} Time is discretised into epochs of length $\Delta t$, and a request expires after $\tau_{\max}$ epochs without assignment. Both values follow the simulator of prior work~\cite{danassis2019putting} and place the system firmly in the high-frequency dispatch regime.

These four assumptions define the static dispatch sub-problem that JointMatch solves at each epoch; the temporal coupling between epochs is handled by the simulator described in \S\ref{sec:experiments}.

\subsection{Hyperparameter values}
\label{sec:appendix-hp}

\begin{table}[h]
\centering
\caption{Numerical settings for all symbols introduced in the main text.}
\label{tab:hyperparams}
\small
\begin{tabular}{l l l}
\toprule
Symbol & Meaning & Value \\
\midrule
\multicolumn{3}{l}{\emph{Simulator (\S\ref{sec:problem-joint}, \S\ref{sec:appendix-assumptions})}} \\
$\Delta t$              & Dispatch epoch length              & 1 minute \\
$\tau_{\max}$           & Request expiration threshold       & 5 epochs (15 min) \\
$C$                     & Vehicle seat capacity              & 4 \\
$\bar{v}$               & Average travel speed (Manhattan)   & 22.3 km/h \\
\midrule
\multicolumn{3}{l}{\emph{Pricing (\S\ref{sec:problem-joint}, after~\cite{danassis2019putting})}} \\
$\beta$                 & Per-request drop fee               & \$2.20 \\
$\alpha_1$              & Solo fare rate                     & \$0.994 / km \\
$\alpha_2$              & Shared (pooled) fare rate          & \$0.800 / km \\
$\gamma$                & Operating cost                     & \$0.069 / km \\
\midrule
\multicolumn{3}{l}{\emph{Sparsified hetero-graph (\S\ref{sec:method-graph})}} \\
$k_v$, $k_r$            & KD-tree neighbourhood sizes        & 16, 16 \\
\midrule
\multicolumn{3}{l}{\emph{Hetero-GNN scorer (\S\ref{sec:method-scorer})}} \\
$H$                     & Hidden dimension                   & 64 \\
$L$                     & Number of message-passing layers   & 6 \\
$\varepsilon$           & Aggregation stability constant     & $10^{-6}$ \\
\midrule
\multicolumn{3}{l}{\emph{One-shot decoder (\S\ref{sec:method-decode})}} \\
$\lambda$               & Pairing-edge weight in plan utility & 0.5 \\
\midrule
\multicolumn{3}{l}{\emph{Stage A — supervised pretraining (\S\ref{sec:training})}} \\
                        & Optimiser                          & Adam \\
                        & Learning rate                      & $5\times 10^{-4}$ \\
                        & Epochs                             & 8 \\
                        & Training fleet size                & 50 \\
                        & Loss                               & BCE-with-logits, class-weighted \\
                        & Pos-class weight (vr / rr)         & $\approx$ 5 / 2 \\
                        & Best validation BCE                & $\approx 0.78$ \\
\midrule
\multicolumn{3}{l}{\emph{Stage B — REINFORCE fine-tuning (\S\ref{sec:training})}} \\
                        & Optimiser                          & Adam \\
                        & Learning rate                      & $10^{-5}$ \\
                        & Episodes                           & 15 \\
                        & Entropy coefficient $\alpha$       & $5\times 10^{-3}$ \\
                        & Baseline                           & moving average ($\rho = 0.9$) \\
                        & Gradient clipping (max norm)       & 1.0 \\
                        & Sampler temperature                & 1.0 \\
\midrule
\multicolumn{3}{l}{\emph{Evaluation protocol (\S\ref{sec:experiments})}} \\
                        & Random seed                        & 42 \\
                        & Days per (method, fleet) cell      & 3 \\
                        & Fleet sizes swept                  & 20, 50, 100, 200 \\
\bottomrule
\end{tabular}
\end{table}

\end{document}